\documentclass[journal]{IEEEtran}
\usepackage{algorithm}

\usepackage{amsmath}
\usepackage{graphicx}
\usepackage{amsmath,amssymb,amsfonts}
\usepackage[noend]{algpseudocode}
\usepackage{color,soul}
\usepackage{subfigure}
\algrenewcommand{\Return}{\State \textbf{return}\ }
\algnewcommand{\funccall}[1]{\textit{#1}}
\ifCLASSINFOpdf
\else
\fi
\begin{document}

%
\title{ A Data Augmented Approach to Transfer Learning for Covid-19 Detection}

%
%
%

\author{Shagufta~Henna,~\IEEEmembership{Senior Member,~IEEE,}
 Aparna Reji

\thanks{Shagufta Henna and Aparna Reji are with the Computing Department, Letterkenny Institute of Technology, Letterkenny, Co. Donegal, Ireland.
e-mail: shaguftahenna@gmail.com

}
\thanks{Under review \newline
Manuscript received XX,XXXX; revised XX, XX.}}

%
%

\markboth{  }%
{Shell \MakeLowercase{\textit{et al.}}: Bare Demo of IEEEtran.cls for IEEE Journals}
%



\maketitle

\begin{abstract}
Covid-19 detection at an early stage can aid in an effective treatment and isolation plan to prevent its spread.  Recently, transfer learning has been used for Covid-19 detection using X-ray, ultrasound, and CT scans. One of the major limitations inherent to these proposed methods is limited labeled dataset size that affects the reliability of Covid-19 diagnosis and disease progression.  In this work, we demonstrate that how we can augment limited X-ray images data by using Contrast limited adaptive histogram equalization (CLAHE) to train the last layer of the pre-trained deep learning models to mitigate the bias of transfer learning for Covid-19 detection. We transfer learned various pre-trained deep learning models including  AlexNet, ZFNet, VGG-16, ResNet-18, and GoogLeNet, and fine-tune the last layer by using CLAHE-augmented dataset. The experiment results reveal that the CLAHE-based augmentation to various pre-trained deep learning models significantly improves the model efficiency. The pre-trained VCG-16 model with CLAHE-based augmented images achieves a sensitivity of 95\% using $15$ epochs.  AlexNet works show good sensitivity when trained on non-augmented data. Other models demonstrate a value of less than $60\%$ when trained on non-augmented data. Our results reveal that the sample bias can negatively impact the performance of transfer learning which is significantly improved by using CLAHE-based augmentation.
 
 \end{abstract}


%
\IEEEpeerreviewmaketitle

\section{Introduction}

\IEEEPARstart{R}ecent years have witnessed the popularity of transfer learning to accelerate deep learning approaches to solve complex classification problems. Transfer learning offers these benefits by using pre-trained models that accelerate deep learning deployment for different computer vision applications including robotics, video processing, and retrieval. Transfer learning exploits the correlation between datasets to devise classification algorithms that can be deployed directly in dynamic environments\cite{van15}. 
In Transfer learning, a base deep neural network (DNN) is trained on a given dataset to extract features that are given as input to the second DNN which is trained on a target dataset \cite{Reddy19}. Conventional image classification techniques rely on image segmentation preceded by segmented object recognition using statistical or machine learning classifiers. These multi-class classification algorithms are computationally expensive and time-consuming \cite{Kermany18}.This challenge has been addressed by the introduction of the Convolutional Neural Network (CNN). CNN improves image classification and object recognition by generating an abstract image representation called a feature map using convolution layers and image filters. One of the major shortcomings of CNN is that it requires a large number of relevant training samples for good results.  Transfer learning addresses this limitation and can work efficiently even with limited training datasets. A transfer learning approach uses a feed-forward neural network coupled with a back-propagation method. In the first instance, the approach fine-tunes the weights of the lower layers to detect general patterns. Secondly, the weights of the upper layers are retrained to recognize unique features inherent to an image \cite{yang18}.

Covid-19 is an infectious disease caused by severe acute respiratory coronavirus syndrome (SARS-CoV-2) \cite{dong20}. The disease was first identified in Wuhan, China, in December 2019, and soon was a pandemic with millions of reported deaths. Early and faster Covid-19 detection can help to contain the disease and can relieve some stress from healthcare.  Some of the widely used  Covid-19 screening techniques include  Reverse transcriptase quantitative polymerase chain reaction (RT-qPCR), computed tomography (CT) imaging, thermal screening, and X-Ray imaging \cite{wang20}. 
These screen techniques have some serious drawbacks. For example, in RT-qPCR test, viral RNA is extracted from nasal swabs to visibly spot the virus using a fluorescent dye that requires human commitment and is time-intensive \cite{chen20}.

Similarly, CT scanners  require substantial cleaning after each use with higher radiation risks to patients. As Covid-19 damages human respiratory cells by flooding alveoli with pus and blood. This results in patchy opacities across the mid-zone, lower or upper zones of lungs which can appear in chest X-rays. This makes the chest X-rays an optimal way to identify Covid-19 than thermal screening \cite{Wong20} compared to other screen methods. Computer vision-assisted diagnostic tools for image classification can be used to assist the clinician with a second opinion for better decision-making to prevent infectious diseases. In this context, various deep learning approaches to medical image classification have been reported in the literature. Recently, Hemdan et al. \cite{Hemdan20} proposed a deep learning model called COVIDX-Net based on seven CNNs to diagnose Covid-19 using X-ray images. Similarly, another work in  \cite{wangl20} has introduced a deep model for Covid-19, called COVID-Net to classify Covid-19 and non-Covid-19 patients. These deep learning approaches, however, assume a large labeled X-ray dataset is available to train the deep learning models which is an expensive and tedious process. Further, due to the fact, the Covid-19 outbreak is recent, it is unrealistic to collect relevant X-ray images to train DNNs \cite{waheed20}. Further, DNNs tend to overfit when there is not enough training data to fit the parameters of each layer, resulting in poor generalization \cite{Heidari20,wang20}.

These challenges can be alleviated by exploiting the potentials of transfer learning that results in high-quality DNN models with small data set \cite{Zhao19}. Further, transfer learning eliminates the DNN training complexity due to a higher number of parameters. In the context of Covid-19, transfer learning trains DNNs on a SARS-CoV-2-related X-ray dataset to initialize DNNs. These pre-trained DNNs are fine-tuned then with minimal Covid-19 dataset. Consequently, transfer learning offers remarkable results in form of fully trained DNNs. The transfer learning techniques, however, tends to create sample biasness which can degrade the reliability of deep neural network approaches.
This biased transfer learning behaviour can be resolved with appropriate data augmentation methods that are capable to extend the training dataset.  Data augmentation has been used widely for image transformation, scaling, flipping, blurring, sharpening and white balance, etc \cite{Mik18}. These augmentation methods are fast and easy, however, these methods result in a slight alteration to training samples which affect can even affect the reliability of transfer learning. On the contrary to these augmentation methods, in this work, we propose a CLAHE(Contrast Limited Adaptive Histogram Equalization) algorithm-based enhancement to transfer learning \cite{Ma19}. The CLAHE locally enhances the low-contrast X-ray images and provides better feature details compared to other augmentation methods. In summary, the contributions of this research paper are as follows:

\begin{itemize}

\item Propose transfer learning models based on various deep learning approaches, i.e., AlexNet, ZFNet, VGG-16, ResNet-18, and GoogLeNet with CLAHE-based augmentation on chest X-ray images.
\item Evaluate the performance of CLAHE-based transfer learning approaches to the above models in terms of sensitivity, specificity, accuracy, and confusion matrices
\item Compare the CLAHE-based transfer learning to the AlexNet, ZFNet, VGG-16, ResNet-18, and GoogLeNet  to the transfer learning without augmentation for sensitivity, MCC, specificity, accuracy, and confusion matrices.
\end{itemize}

The rest of the article is organized as follows. Section \ref{rw} discusses the related works for transfer learning for Covid-19 detection. Section \ref{clahe} discusses the CLAHE-enabled augmentation which is used to transfer learn the models in Section \ref{clahemodels}. Our experimental setup, results, and analysis are presented in Section \ref{eval}. Finally, Section \ref{conclude} concludes the paper with future research directions.

%
%
%
%

 
\section{Related Works} \label{rw}

In the last few months, various machine learning approaches including deep learning to diagnose Covid-19 have been proposed \cite{Haghanifar20,Barstugan20}. Authors in \cite{wang20} have developed a model called COVID-Net to identify anomalies in chest X-ray images of healthy individuals and patients suffering from viral pneumonia, bacterial pneumonia, and Covid-19. COVID-Net is the first opensource application to detect Covid-19 images from the X-ray chest. The COVID-Net architecture consists of a mixture of diverse convolution layers with filter size varying from 7 X 7 to 1 X 1 and grouping configurations. The model used 22 epochs to train with a batch size of 64 and the Adam optimizer. COVID-Net obtained an accuracy of 92.6\% with a sensitivity of 87.1\%, 97.0\%, and 90.0\%  for Covid-19, normal, and pneumonia classes, respectively. One of the major issues with the COVID-Net is that it does not generalize well due to the small data size.

Authors in \cite{Narin20} proposed a model based on InceptionV3, ResNet50 and InceptionResNetV2 for Covid-19 detection from chest X-rays using 5-fold cross-validation. The model is trained using $50$ Covid-19 and $50$ normal X-ray images for 30 epochs. Among all models, ResNet50 shows the highest sensitivity of 96\%  with less training time. However,   the initial achieved results are below 70\% due to limitations on dataset size.

In \cite{Hall20}, authors have proposed a model for Covid-19detection from chest X-ray images consisting of Covid-19, viral, and bacterial pneumonia. The model follows the transfer learning approach by fine-tuning Resnet50 and VGG16 on a small dataset with horizontal flip augmentation. This approach uses 10-fold cross-validation to produce 21 models and considers average accuracy over all the test predictions. The model reports 80.39\% accuracy for Covid-19 detection. One of the shortcomings of this approach is insufficient chest X-ray images augmented with low-resolution Covid-19 X-ray images.

Authors in \cite{Cohen20} implemented a pre-trained DenseNet model to extract features based on X-ray images.  The proposal model predicts the severity of Covid-19 for efficient treatment in ICU. To improve training performance, the proposed model considers augmentation by cropping, re-sizing, and scaling with a mean squared error of 0.78 for lung opacity score. The proposed model immune to over-fitting, however, it is complex to implement. Further, the DenseNet model pre-training requires large X-ray images dataset.

Authors in \cite{Zhong20} proposed a deep CNN model to screen Covid-19 using CXR Dataset with 6,354 chest X-ray images. The model accelerates chest X-ray scanning for Covid-19 detection. The model enhances the quality of images by horizontal flip, brightness adjustment, cropping, and contrast adjustment. The Deep CNN model is fine-tuned by applying a Global-Average-Pooling layer to VGG-16. The model demonstrates the sensitivity of 84.46\% when tested. Nevertheless, the images obtained from various sources are not validated in consultation with a radiologist. In a work \cite{Khan20} proposed a model called CoroNet that used CNN and pre-trained on Xception model to detect Covid-19 from chest X-ray images. The model demonstrated an accuracy of 89.6\ with a sensitivity of  93\%.  The model uses 4-fold cross-validation to evaluate model output from pneumonia bacterial, viral, normal, and Covid-19 classes.

In \cite{Rajaraman20} used transfer learning approach based on pre-trained  CNN and ImageNet models to extract features for COVD-19 detection. Authors, further, have applied recursive pruning using 
 Inception-V3  to reduce the complexity of the approach. The model demonstrated a score of 0.99 to detect Covid-19 from pneumonia and normal images. Despite these benefits, the proposed model still remains complex with a high number of hyperparameters. Luz et al. \cite{Luz20} implemented an EfficientNet model for  Covid-19 detection trained on a dataset of 13,569 X-ray images. The X-ray images are enhanced scaling, rotation, and horizontal turn.  The model is fine-tuned with fully connected layers with batch normalization, swish activation, and drop-out functions. The model demonstrated an accuracy of 93.9\% with a  sensitivity of 96.8\%. The work also considers other pre-trained models for transfer learning including VCG, MobileNet, and ResNet.

Apostolopoulos et al. \cite{Apostolopoulos20} propose VGG16, Inception, Inception ResNet v2, Mobile Net, and Xception models for Covid-19 detection. The work considers a dataset with  500 normal, 224 Covid-19, and 700 bacterial pneumonia chest X-rays. The VGG-16 model shows the highest accuracy of 98.78\% in binary classification and an accuracy of 93.48\% in multi-class classification. In \cite{Boudrioua20}, authors propose various approaches to transfer learning for early diagnosis of Covid-19 from chest X-ray images. Authors have evaluated the performance of  NASNetLarge, DenseNet 121, and NASNetMobile; three pre-trained CNNs that are fine-tuned for training. The dataset used for this study consists of 309 Covid-19, 2000 pneumonia, and 1000 healthy chest X-ray images. NASNetLarge shows an average sensitivity of 99.45\% with a specificity of 99.5\%.  One of the major shortcomings of this work is an imbalanced dataset with no augmentation.

In a work in \cite{Tabik20}, authors proposed Covid-19 SmartData based network (COVID-SDNet) model based on a large dataset of chest X-rays called COVIDGR-1.0 for early Covid-19 detection.  A pre-trained Resnet-50 is transfer learned to build the COVID-SDNet. The model achieves an accuracy of $66.5\%$, $88.14\%$, and $97.37\%$ in mild, moderate, and severe Covid-19 intensity levels. Although the model performs better, however, it is not reliable as the predictions are derived from a single model. A work in \cite{Bassi20}, authors have proposed a two-fold transfer learning approach for Covid-19 diagnosis from chest X-ray images. The proposed approach consists of a three-step learning process, i.e., selection of DenseNet201, pre-trained on ImageNet, and retraining it on NIH ChestX-ray14 dataset.  The model obtains a test accuracy of $99.4\%$ and an F1 score of 0.994. Initially, the model provides a good classification, however, its performance does not improve with an increase in the number of epochs. In a work in \cite{Hemdan20}, authors developed a deep learning framework called COVIDX-Net for the identification of Covid-19 from X-ray Images.  The model is based on seven deep CNNs namely DenseNet201, InceptionV3, Xception, ResNetV2, MobileNetV2, VGG19 and InceptionResNetV2. The algorithms are trained on a mini dataset of 50 X-ray images with 25 normal images and 25 positive Covid-19 images. The VGG19 and DenseNet201 models showed reliable results in the Covid-19 classification with F1-scores of 0.89 and 0.91. The proposed architecture is deeper with a high number of layers and thus is more costly to train. 

In another work \cite{abbas20}, authors have proposed a model called DeTraC that uses the ResNet-18 model paired with the class decomposition technique for Covid-19 detection based on chest X-ray images. The model aims to address irregularities such as overlapping classes present in images by finding class boundaries with class decomposition techniques. The transfer learning approach uses an AlexNet model to extract features. The ResNet-18 coupled with DeTraC model achieves a sensitivity of 97.91\% with an overall accuracy of 95.12\%. Conversely, the model uses 256 epochs for training and thus results in increased training time.

In authors \cite{basu20} proposed a method called Domain Extension Transfer Learning (DETL) that uses deep CNN for the classification of pneumonia, normal, Covid-19, and other diseases from chest X-ray dataset. The model uses real-time augmentation like flipping, scaling, and mirroring to the dataset prior to training  The overall accuracy of the proposed model is 95.3\%. The model is trained for $100$  iterations which incur high computational processing.Authors in \cite{Farooq20} proposed a model called COVID-ResNet based on ResNet-50 for  Covid-19 detection from chest X-rays. The model is trained using 5941 X-ray images consisting of Covid-19, viral pneumonia, bacterial pneumonia, and healthy individuals. Dataset is augmented by using simple enhancement techniques such as vertical flip, random rotations, and lighting conditions to improve the generalization of the model. The proposed model demonstrates a sensitivity of 100 \% for the detection of Covid-19.  The model consists of 50 layers that increase the need for more computational resources and training time.

Authors in \cite{Punn20} proposed a fine-tuned transfer learning of Inception-v3, Inception ResNet-v2, NASNetLarge, DenseNet169, and ResNet coupled with random oversampling and weighted class loss function for classifying Covid-19 and normal classes. The proposal addresses the issue of unbiased learning caused by insufficient samples for the training of Covid-19 detection models. In the weighted class technique, to balance the data, other classes with low training samples are configured to carry more weights during training.  Random sampling is achieved with the help of rotation, scaling, and displacement. Among all the transfer learning models, NASNetLarge achieves a result with 0.91 sensitivity in binary class classification. 

In a work in \cite{Minaee20}, authors used various pre-trained models including SqueezeNet, DenseNet-121, ResNet50, and ResNet18 for faster Covid-19 diagnosis. The model is trained on a dataset of 071 X-ray images augmented using standard augmentation techniques to overcome the dataset imbalance. Almost all model achieves a sensitivity of 98\% with an accuracy of 90\%. Authors in \cite{waheed20} introduced a model called CovidGAN based on Auxiliary Classifier Generative Adversarial Network (ACGAN) to generate chest X-ray images to be used in Covid-19 detection. The model was developed to cope with constraints of Covid-19 positive radiographic images dataset. A VGG-16 network with four custom layers which includes a global average pooling layer, 64 units dense layer, dropout layer, and a softmax layer. The model achieves a sensitivity of 69\% when trained on real data and a sensitivity of 90\%when trained on augmented data. The ACGAN is a variant of Generative Adversarial Networks (GANs) based on two neural networks that require high computational capabilities. Also, the dataset is collected from various sources and is not cross-validated which can affect model accuracy. 

A dominant challenge in medical image analysis is limited data availability \cite{Shorten19}. Data augmentation augments the data in terms of quantity and diversity. In data augmentation, new samples are created by applying small mutations to the original data. It can generate massive data that minimizes over-fitting on deep learning models, thereby improving versatility and generalization \cite{Hu20}. Augmentation can emulate various features in images to improve the performance of deep learning-based classification in terms of accuracy.Some basic image augmentation techniques include flipping, rotation, cropping, and translation \cite{McGuinness20}. Flipping is one of the easiest and useful augmentation techniques. In flipping,  images are flipped on the horizontal axis instead of the vertical axis. Cropping is another augmentation technique used in images with mixed height and width dimensions to create a focal fix for each image. The translation move images left, right, up, and down. In an image translation, space is filled with a constant value, i.e., 0 or 255, or with random or Gaussian noise.  The image cropping reduces the size of the image, whereas image translation retains the spatial dimensions of images. Rotation augmentation pivots an image left or right on an axis in the range of 1 and 359 degree \cite{Shorten19}.

A comprehensive review and investigation of transfer learning techniques for Covid-19 detection reveal that most of these techniques suffer in terms of the imbalanced dataset. Some transfer learning techniques that rely on a single model result in high bias. Further, the training of deep learning techniques requires a large dataset to learn from general to specific patterns with a higher number of parameters that should be tuned, thereby resulting in a complex training process. To resolve the above challenges, in this work, we use a transfer learning method for various pre-trained DNNs with less number of hyperparameters and low computational resource requirements. Further, to address the issue of bias in training samples to fine-tune the different pre-trained models, we propose a CLAHE-based augmentation method for Covid-19 detection. To our best knowledge, this is the first work to augment X-ray images data using CLAHE-based augmentation to realize transfer learning.

\section{CLAHE-augmented Transfer Learning for Covid-19 Detection} \label{clahe}

\subsection{CLAHE for Data Augmentation}
CLAHE is a contrast enhancement algorithm that improves the image quality by highlighting the image features and information. Image contrast is enriched by local or global pixel processing. Some traditional contrast enhancement algorithms, e.g., Histogram Equalization (HE), adaptive histogram equalization (AHE), and hybrid cumulative histogram equalization (HCHE) redistribute the brightness of an image using histograms \cite{Ting2016}.

\begin{figure}[ht] 
    \includegraphics[width=.5\textwidth]{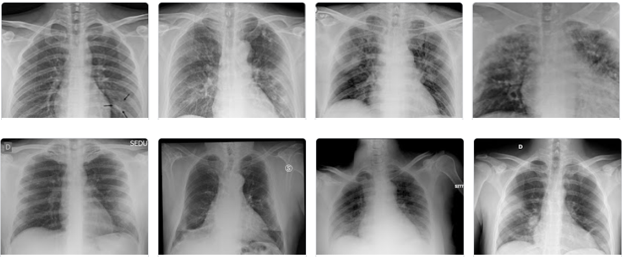}   
    \caption{Non-augmented chest X-ray images.} \label{nonaugimg} 
\end{figure}

\begin{figure}[ht] 
    \includegraphics[width=.5\textwidth]{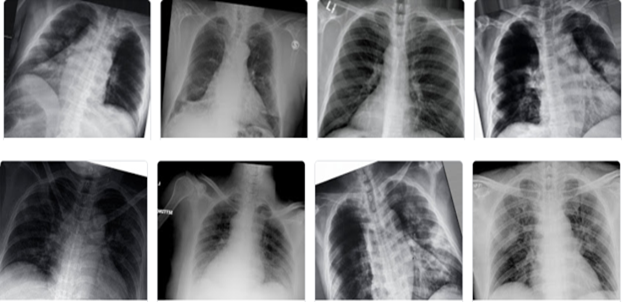}   
    \caption{CLAHE-enabled augmented chest X-ray images.} \label{augimg} 
\end{figure}

CLAHE algorithm overcomes the issue of global histogram equalization incurred due to noise amplification in homogeneous regions. It computes different histograms and utilizes them to redistribute the lightness value \cite{Ma19}.  Primarily, CLAHE enhances the low-contrast medical images with the help of control image enhancement quality parameters, i.e.,  Block Size (BS) and Clip Limit (CL) \cite{Koonsanit2017}. CL limits the noise amplification by clipping histograms at a specified value before computing the Cumulative Distribution Function (CDF). Figure \ref{nonaugimg} and Figure \ref{augimg} show non-aumgneted and augmented chest X-ray images.

Algorithm \ref{algo1} illustrates the steps involved in data augmentation to address the issue of an imbalanced dataset of lung X-ray images. Line 1 in Algorithm \ref{algo1} shows that a labeled Algorithm \ref{algo1} takes the Covid-19 dataset as an input. Line 2 shows that the Algorithm returns the output of augmented Covid-19 images.   Line 6 to 11 gives the steps to perform augmentation on the Covid-19 dataset.  Line 7 outlines the steps of selecting random images and applying a 20 degree rotation using ImageDataGenerator(). A random-rotation with a limit of 25 degree ensures that the portraying class features are not removed by cropping \cite{McGuinness20}. Line 8 and line 9 demonstrates the method of translating the images both horizontally and vertically. Line 10 describes the process to flip random images to generate new. It also applies the default likelihood of flipping to each image. Line 11 presents the process of filling points residing outside the boundary of the input image using the nearest points.

Finally, line 12 shows the  CLAHE-based augmentation by calling the AHE() method.  Line 15 illustrates the process to add a CL of 0.03 to the images using the CLAHE.  CLAHE divides an image into several non-overlapping regions of equivalent size.  It determines histograms for each region that are redistributed such their height do not exceed 0.03. CLAHE generates multiple images after applying the contrast enhancement mechanism. The unevenness in the dataset is sorted before applying augmentation. This results in an increase in the dataset with 1847 Covid-19 labeled images that are used for training. 

\begin{algorithm}
  Input Data: Chest X-ray Images (X,Y); $Y ={y/y \in {Covid-19}}$ \\
   Ouput Data: (X,Y); where $Y={Y \in {Covid-19_{Aug}}} $ 
  \Comment{The augmented Chest X-ray Image Dataset}
  \vspace{2.5mm}
  \begin{algorithmic} [1]

    \Loop 
      \State \funccall{for all $x \in X$} 
           \Comment{Apply ImageDataGenerator to all images}
         \State $x \leftarrow  ImageDataGenerator(x)$  
            \EndLoop    
            \vspace{2.5mm}
           \State \textbf {function \funccall { \textbf {ImageDataGenerator}}} (x)
            \State  \hspace{3mm}  rotate( random( x,20))
           \Comment{Rotate images at 20} 
             \State  \hspace{3mm} shift(width, 0.01)
           \Comment{Shift width to 0.01} 
                  \State  \hspace{3mm} shift(height,0.01)
           \Comment{Shift height to 0.01} 
                \State   \hspace{3mm} flip(random(x, horizontal)
           \Comment{Flip horizontally} 
                \State   \hspace{3mm} fill(mode=nearest)
           \Comment{Fill points outside boundaries}      
           \State  \hspace{3mm} $x \leftarrow  AHE(x)$  
            \Comment{Apply AHE to input images}
           \State  \textbf{end ImageDataGenerator()} 
                       \vspace{2.5mm}   
            \State \textbf {function \funccall {AHE}} (x)
            \State   CLAHE(clip\_limit=0.03) 
            \Comment{Apply CLAHE algorithm with contrast enhancement}
             \State  \textbf{end AHE()}

  \end{algorithmic}
  \caption{CLAHE-based Image Augmentation Algorithm \label{algo1}}
\end{algorithm}

\section{ Transfer Learned Models based on CLAHE for Covid-19 Detection} \label{clahemodels}
The models used for transfer learning based on CLAHE-based augmentation are discussed below:

\subsection {Selected Models}

\subsubsection{AlextNet}
AlexNet is a CNN model that won the ImageNet Large Scale Visual Recognition (ILSVRC) for detecting objects and classifying images in 2012 \cite{Russakovsky15}. The model is based on five convolutional layers with three fully-connected layers. The first two convolutional layers of the model are backed by normalization and max-pooling layers that use a $3 \times 3$ filter and a stride size of 2. The next two convolutional layers are linked straight with a max-pooling layer followed by the last convolutional layer in the model. The output of the max-pooling layer is passed to two fully connected layers, of which the second fully connected layers that are supported by a softmax classifier. The filters present in the first and second convolutional layers are of size $11 \times 11$ and $5 \times 5$, respectively. A regularization method called dropout using a 0.5 ratio is used to avoid over-fitting in fully connected layers. After each of the first seven layers, the Rectified Linear Unit (ReLU) is used to train nonlinearity faster \cite{Maeda20}.

\subsubsection{ZFNeT}
ZFNet is a CNN architecture built as an extension to AlexNet. The model uses $7 \times 7$ kernels rather than $11 \times 11$  to reduce the number of weights. The model aims to prevent pixel information due to the large filter and uses ReLu as the activation function \cite{Gothai20}.  It not only reduces the number of network parameters but also increases the accuracy of the classification.

\subsubsection{VGG-16}
VGG-16 is a deep CNN that consists of sixteen layers. The architecture of this model involves five stacks of convolutions. Spatial resolutions are retained after $3 \times 3$ convolution layers with the aid of 1-pixel padding and 1-pixel stride. There are five max-pooling layers with stride 2 that trail some convolutional layers.  The classifier part consists of three fully connected layers, the first two layers having 4096 channels and the third layer having as many channels as the number of classes. The third fully connected layer that performs the classification is followed by a final soft-max layer that yields class probabilities \cite{Alshalali18}. VGG-16 is an improvement to AlexNet model by supplanting large kernel-sized filters such as 11 and 5 in the first and second layers of convolution with multiple $3 \times 3$ kernel-sized filters.

\subsubsection{ResNet-18}
The architecture of ResNet is renowned for its use of residual blocks. This model architecture consists of a continuous connection between residual blocks to address gradient disappearance issues caused by the stacking of deeper layers in CNN \cite{Simon19}. Residual block is a fundamental concept that attaches input x to output $F(x)$ to prevent gradient disappearance while using a $3 \times 3$ convolution layer twice \cite{Oh2019}. The model comprises eight modules, each of which has two convolutional layers where each filter is accompanied by batch normalization \cite{Alinsaif20}.

\subsubsection{GoogLeNet}
GoogLeNet is a model that comprises 22 layers proposed to reduce the computational complexity associated with traditional CNN. This model is the winner of the 2014 ILSVRC challenge \cite{Szegedy20}. The model uses fewer network parameters than AlexNet and VGG. The intricacy of the model is reduced by incorporating inception layers based on different kernel sizes. These kernels perform dimensionality reduction to avoid computationally expensive layers. 

\subsection{Proposed Transfer Learned Models for Covid-19 Detection}

All the selected models for Covid-19 detection are pre-trained on ImageNet dataset that consists of  1.2 million high-resolution images from 1000 classes. These pre-trained models cannot directly extract features from the ImageNet dataset to classify the images. To address this challenge, these models are fine-tuned on X-ray image dataset to extract Covid-19 features. During fine-tuning, the output produced by the fully connected layer of all the selected models is modified from $1000$ to $2$ with binary classification performed by the softmax layer.

Using a higher value as a learning rate can result in the neural network to start diverging instead of converging. A lower learning rate can take longer for the neural network to converge with a chance to lost in local minima. This can be addressed by reducing the learning rate during training. The initial learning rate of all the selected models is set to 0.01 and is decreased to 0.01 after every 4 epochs. 
Batch-size is another important parameter that can affect the output of a deep learning model. It helps to converge faster to reach a better minima during model training. In model training, the use of a larger batch size decreases the number of times the entire model undergoes parameter synchronization. Nevertheless, the use of very large batch sizes can negatively affect the performance of the algorithm due to higher convergence. The batch size of each selected model for transfer learning is set to 16.

The loss function measures the CNN predictions on a training set. Since the learning rate has a direct impact on the model, learning rate algorithms or optimizers are used to set the learning rate based on the loss function to boost the model. Optimizers, e.g., Gradient descent aims to reduce the loss function.  Gradient descent minimizes the loss function by updating the weights and bias \cite{Pomerat19}. Other examples of optimizers include RMSProp algorithm, Adam algorithm, gradient descent algorithm, FTRL algorithm, and Adagrad algorithm \cite{Pomerat19}. Adam optimizer is a combination of RMSprop and Stochastic Gradient Descent with the momentum that employs squared gradients like RMSprop to scale the learning rate \cite{Pomerat19}. We have selected Adam optimizer as the loss function that converges faster and uses less number of iterations to reach the local minima of the loss function \cite{Pomerat19}.

Algorithm \ref{Algo2} explains the proposed transfer learning approach to detect Covid-19 from chest X-ray images.  Line 1 shows that all images from the dataset are taken as input. Line 2 shows the expected output where the images in $x \in X$ are correctly labeled as$y \in Y$. Line 1 shows a call to Algorithm \ref{algo1} to perform data augmentation on images labelled as Covid-19. Line 2 executes the loop until there are images in the dataset.  Line 3, further modifies the image to $224 \times 224$ before feeding it to the pre-trained models.  Line 4 illustrates the method of choosing random images from the resized dataset to rotate at an angle of 20 degree. Line 5 represents the method to flip the images horizontally, where the probability to select an image for flipping is set to 0.5. Line 6 shows the conversion of images to tensors. Line 7 normalizes the image using the mean and the standard deviation. Line 8  downloads various pre-trained models D for transfer learning including AlexNet, ZFNet, VGG-16, ResNet-18, and GoogLeNet.

Line 10 calls fit () mechanism where the models are fine-tuned, and transfer learned. Line 17 demonstrates the method of fit () replacing the last layers of all models with a classifier that outputs two classes. Line 16 to 17 shows the process of applying a learning rate of 0.01 for each model. Line 19 demonstrates the method of training these fine-tuned models for 5, 10, 15, 20, and 25 epochs. Line 20 indicates that, for each training with a batch size 16, the model parameters are updated during backpropagation and after four training epochs.  Line 23 shows that if the loss increases, the models are set to learning by 0.01.  Line 11 to 12  show the process of loading test data $x \in X_{test}$  to make predictions by using a fine-tuned model from D.
 
\begin{algorithm}
  Input Data: Chest X-ray Images (X,Y); $Y ={y/y \in {Covid-19, normal}}$ \\
   Ouput:  $\forall d \in D \{( Y_{i}) \in (Y_{test})\}=Y$
  \Comment{Transfer learned models for Covid-19 detection}
  \vspace{2.5mm}
  \begin{algorithmic} [1]
               
     \State   $X  \leftarrow \funccall { Algorithm 1(X)}$
      \Comment {Generate augmented images by using Algorithm \ref{algo1} }
     \While {$X != \emptyset$}

      \State  resize(x, $224 \times 224$) 
      \Comment {Modify image dimension }
      \State  rotate( random( x,20))
      \Comment{Rotate images at 20}       
      \State   flip{(x, horizontal):$P(r(x))=0.5$}
      \Comment{Flip horizontally with probability of rotation 0.5 }
      \State   $x \leftarrow  totensor(x)$  
   \Comment{Convert image to tensor}
      \State   $x \leftarrow {\bar{x}(0.485, 0.456, 0.406) \& std(0.229, 0.224, 0.225)}$ 
     \Comment{Normalize each image }
        \State   $D \leftarrow$ \{AlexNet, ZFNet, VGG-16, ResNet-18, GoogLeNet\} 
   \Comment{Download and fine-tune models}
      \vspace{2.5mm}
  \For {all $d \in D$}
     \State   $d \leftarrow fit(d)$  
     \State  \textbf {foreach} $x \in X_{test}$ 
         \State  d.predict\_Covid-19(x)
    \EndFor
          
     \EndWhile             
          \vspace{2.5mm}
           \State \textbf {function \funccall {fit}} (d)
           \State  replace(classifier($d(2,dim=1))$)
           \Comment{Replace last layer of each pre-trained model} 
            \State $\mu \leftarrow  0.01$
            \State  \textbf{end fit()} 
                       \vspace{2.5mm}

 \For {$epoch \leq 25$}
     \State   $d \leftarrow fit(d)$  
     \State  \textbf {foreach} $batch (X_{i}, Y_{i}) \in (X_{train}, Y_{train})$ 
         \State update(d, param) 
            \Comment{Update parameters of model}
           \State $epoch \leftarrow epoch+5$  
           \If {(loss(epochs,4) increases)}
           
             \State $\mu \leftarrow \times \times 0.01$
             \Comment{if loss increases for 4 epochs}
             \EndIf
    \EndFor

  \end{algorithmic}
  \caption{CLAHE-enabled Transfer Learning \label{Algo2}}
\end{algorithm}

\section{Evaluation} \label{eval}
\subsection{Environment Setup}
The process for Covid-19 detection based on transfer learning from pre-trained models require higher computational capabilities. All the pre-trained models including AlexNet, ZFNet, VGG-16, ResNet-18, and GoogLeNet are fine-tuned and re-trained using the target dataset for epochs varying from epoch $5$ to $25$ using Google Colab \cite{Bisong19}. Google Colab provides a free Nvidia Tesla T4 Graphical Processing Unit (GPU) with revolutionary performance. 

For the development of Covid-19 detection models, we use PyTorch \cite{Paszke19}. PyTorch is a well-known Python-based deep learning framework utilized for training and building neural networks \cite{Paszke19}.  Scikit-learn \cite{Pedregosa18}, math, seaborn \cite{Bisong19b}, and matplotlib \cite{Bisong19b} are the other libraries used for the creation of confusion matrix, evaluation metrics calculations, and plotting various graphs.

\subsection{Dataset}
The dataset used for this work is available from the public GitHub repository \cite{Minaee20}. The dataset combines the X-ray images from two other datasets, i.e.,  COVID-Chestxray-Dataset published by Joseph Paul Cohen \cite{Cohen20b} and ChexPert dataset \cite{Irvin19}, a large public dataset for chest radiology images.  The combined dataset consists of 2,031 images for training and 3,040 images for testing. Finally, the resultant dataset is merged with the positive Covid-19 positive images after consultation with a board-certified radiologist. Further, the final dataset is divided into tests and train sets. The training dataset consists of 31 Covid-19 positive images and 2000 non-Covid-19 images. The test dataset consists of 40 Covid-19 and 3000 non-Covid-19 images.  The resultant dataset is highly imbalanced with a limited number of positive Covid-19 images, therefore, it is further augmented to increase the dataset size using CLAHE. The augmented dataset consists of total of 6,887 images, out of which 3,847 images are used for training, and 3,040 images for testing. The training data set includes 1,847 Covid-19 images and 2,000 non-Covid-19 images and the testing dataset with 40 Covid-19 images and 3,000 non-Covid-19 images.  

\subsection{Performance Evaluation Metrics}
The transfer learning models for Covid-19 detection are experimentally analyzed and compared in terms of the following performance metrics.
\subsubsection{Confusion Matrix}
A confusion matrix, also known as a contingency table is a statistical tool used to analyze paired observations \cite{Ariza18}. In this matrix, rows represent an anticipated class and columns a real class, or vice-versa.  For example, in a normal $2 \times 2$ confusion matrix, four integers are listed; 1) true positive (TP) which denotes the number of predicted positive, given it is truly positive), 2) false negative (FN) that is the number of predicted negatives, given it is positive), 3) false positive (FP) representing the number of predicted positives, given it is negative), and 4) true negative (TN) which are the number of predicted negative given it is  negative.

\subsubsection{Sensitivity}
Sensitivity or recall is an indicator of the number of images predicted as positive out of the total number of positive images. It is also referred to as true positive rate (TPR) \cite{Arora16} and is given in Equation \ref{eq1}.

\begin{equation} \label{eq1}
Sensitvity= \small {\frac{TP}{TP+FN}}
\end{equation}

\subsubsection{Specificity}
The specificity or True negative rate (TNR) determines the number of images predicted as negative out of the total images that are negative \cite{Arora16}. It is calculated using Equation \ref{eq2}.

\begin{equation} \label{eq2}
Specificity= \small {\frac{TN}{FP+TN}}
\end{equation}

\subsubsection{Accuracy}
Accuracy characterizes the general correctness of the proposed transfer learning models.  It is measured as the number of correct predictions by the transfer learning model. The accuracy is calculated using Equation \ref{eq3}.

\begin{equation} \label{eq3}
Accuracy= \small {\frac{TP+TN}{FP+TN+TP+FN}}
\end{equation}
\subsubsection{MCC}
Mathews Coefficient Correlation (MCC) measures the performance of a classifier on an imbalanced dataset.  MCC value ranges between -1 and 1, where a value close to 1 implies a good performance and a value closer to -1 indicates the week performance of the classifier \cite{Wardhani2019}. MCC for our proposed transfer learning models is given in Equation \ref{eq4} below.

   \begin{equation} \label{eq4}
MCC= \small {\frac{((TP \times TN)-(FP \times FN))}{\sqrt{(TP+FP)(TP+FN)(TN+FP)(TN+FN)}}}
\end{equation}

\subsection{Performance Analysis of CLAHE-augmented Transfer learning}

In this section, we analyze the effect of hyper-parameters, i.e., learning rate, batch size, and epochs on all the transfer learned models presented in Algorithm \ref{Algo2} with and without augmentation for Covid-19 detection.  Each of these models is trained five-times by fine-tuning the number of epochs to five, ten, fifteen, twenty, and twenty-five. First transfer learned models are trained and tested on the CLAHE-augmented dataset. All the images are down-sampled to $224 \times 224$ before feeding to pre-trained selected models to meet their resolution requirements \cite{Alshalali18}.

\begin{figure}[ht] 
    \includegraphics[width=.45\textwidth]{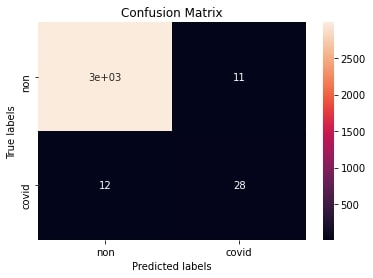}   
    \caption{ResNet-18 with CLAHE-enabled augmentation.} \label{ag-cmresnet-18} 
\end{figure}

Figure \ref{ag-cmresnet-18} represents the confusion matrix obtained while testing the ResNet-18 model. Out of the $40$ Covid-19 positive images, 28 images are categorized as TPs, and $12$ as FNs. This forecast seems sensible in the sense that more than $50$ fifty levels of the all-out positive cases were anticipated to be valid, and just $11$ images are wrongly classified as Covid-19 positive.

\begin{figure}[ht] 
    \includegraphics[width=.45\textwidth]{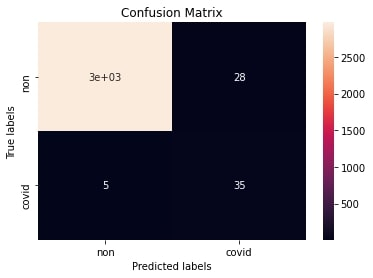}   
    \caption{VGG-16 with CLAHE-enabled augmentation.} \label{ag-vcg-16}  
\end{figure}

The confusion matrix based on the testing of the VGG-16 model trained for 10 epochs is illustrated in Figure \ref{ag-vcg-16}. The model demonstrates excellent performance for Covid-19 detection by characterizing 35 out of 40 images as true positive and misclassifies only 28 images as positive. The model also results in a low FN estimate of $5$ that reduces the risk of undetected Covid-19 patients.

\begin{figure}[ht] 
    \includegraphics[width=.45\textwidth]{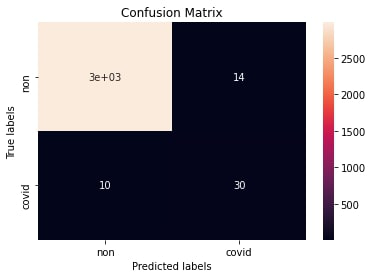}   
    \caption{ZFNet with CLAHE-enabled augmentation.} \label{ag-cf-zfnet} 
\end{figure}

Figure \ref{ag-cf-zfnet} presents the confusion matrix of ZFNet model trained using $20$ epochs.  The model performance is relatively acceptable with $30$ TPs and $10$ FNs. Similarly, only $14$ images have been classified as FPs.  The classification achieved for Covid-19 positive classes is approximately 90\% that makes the model a good fit for Covid-19 detection.

\begin{figure}[ht] 
    \includegraphics[width=.45\textwidth]{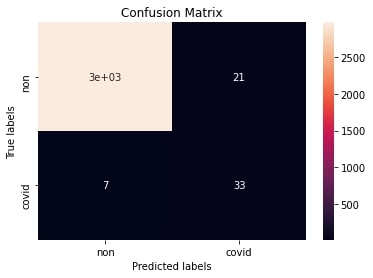}   
    \caption{AlexNet with CLAHE-enabled augmentation.} \label{ag-cf-alexnet} 
\end{figure}

\begin{figure}[ht] 
    \includegraphics[width=.45\textwidth]{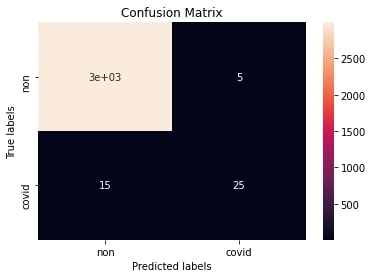}   
    \caption{GoogleNet with CLAHE-enabled augmentation.} \label{ag-googlenet} 
\end{figure}

The confusion matrix for the AlexNet model trained on $10$ epochs is shown in Figure \ref{ag-cf-alexnet}. AlexNet demonstrates good performance to classify the Covid-19 images with 33\% TPs out of $40$ Covid-19. It also lists only $21$  images as FP that appears reasonable to classify TNs correctly from the 2979 images. Figure \ref{ag-googlenet} depicts the confusion matrix for GoogleNet based on $14$ epochs. The model successfully classifies $25$ images as TPs out of the $40$ Covid-19 positive images. This indicates a forecast of over $50\%$ with only $5$ images as FP of healthy patients.

\begin{figure}[ht] 
    \includegraphics[width=.5\textwidth]{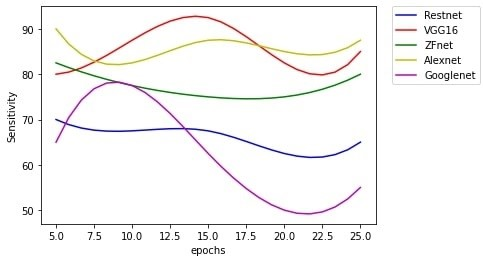}   
    \caption{Sensitivity with augmentation.} \label{senseaugment} 
\end{figure}

In Covid-19 detection, sensitivity plays a vital role as it adequately classifies the Covid-19 positive class from the total Covid-19 positive images fed into the model. Figure \ref{senseaugment} shows the sensitivity of each model trained at different epochs. AlexNet demonstrates a high sensitivity of $90\%$ when trained on $5$ epochs and ZFNet achieves a sensitivity of $82.5\%$. Both models show a decrease in the rate of sensitivity when trained with a higher number of epochs. AlexNet model\'s sensitivity fluctuates with a further ascent in sensitivity level at $15$ and $25$ epochs, and ZFNet model\'s sensitivity is only increased with $25$ epochs. VGG-16 achieves the highest degree with an estimation of $92.5\%$ when trained at $15$ epochs. From $14$ onward, the model\'s performance diminishes with a slight ascent in $25$ epochs. ResNet-18 shows weak performance in terms of sensitivity with a maximum of $70\%$ when trained with $5$ epochs. Among all the models, GoogleNet demonstrates the weakest sensitivity rate of $50\%$ when trained on $25$ epochs. GoogleNet shows the lowest sensitivity value of $77.5\%$ sensitivity.

\begin{figure}[ht] 
    \includegraphics[width=.5\textwidth]{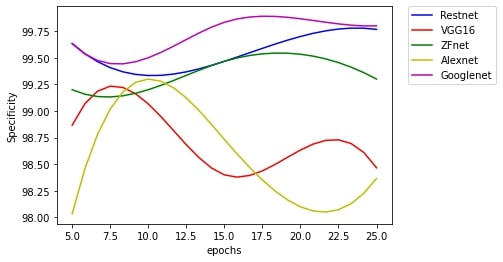}   
    \caption{Specificity  with augmentation.} \label{augspecificity} 
\end{figure}

Figure \ref{augspecificity} delineates the TNR of the $5$ models used for Covid-19 detection trained on augmented data. Both GoogleNet and ResNet-18 show $99.63\%$ specificity when trained on $5$ epochs. 
These models show a marginal degradation in performance when trained on $10$ epochs. However, the model demonstrates an increase in performance with an increase in the number of epochs.  GoogleNet achieves the highest specificity of $99.87\%$ when trained on $20$ epochs. The specificity value of the GoogleNet and ResNet-18 is almost the same when trained on $25$ epochs. ZFNet presents an average performance in terms of specificity with $99.53\%$ being the highest when trained using $25$ epochs. TheVGG-16 demonstrates the least performance while  AlexNet with $98\%$ when trained on $5$ epochs. The model, however, improves its performance to $99\%$ on $10$ epochs. VGG-16 shows a maximum specificity of $99.07\%$ when trained on $10$ epochs.

\begin{figure}[ht] 
    \includegraphics[width=.5\textwidth]{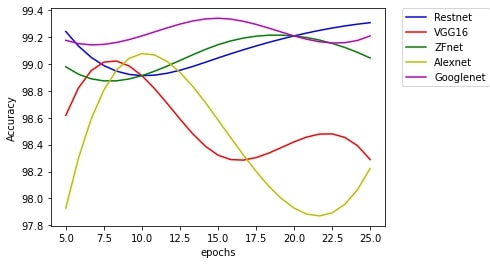}   
    \caption{Accuracy with augmentation.} \label{acc-augment} 
\end{figure}

Figure \ref{acc-augment} shows the accuracy of all the models trained on augmented data. Among all the models, the GoogleNet demonstrates the highest accuracy of $99.34\%$ when trained on $15$ epochs. On the other hand, the AlextNet and VCG-16 when trained on $5$ epochs demonstrate the accuracy of $97.93\%$, $98.62\%$, respectively. However, the accuracy of both the models improves to $99.08\%$ and $98.91\%$ when trained on $10$ epochs. The ResNet-18 model shows an accuracy of  $99.24\%$ for $5$ epochs. ZFNet shows an average performance with $99.21\%$ accuracy when trained on $20$ epochs.

We assess the quality of classification using MCC  on a strongly imbalanced dataset. Figure 12 displays the MCC estimations of different models. It can be observed from the Figure that the GoogleNet achieves highest MCC score of 0.72 when trained on $10$ and $15$ epochs. ZFNet achieves an MCC value of $0.71$ when trained on $20$ epochs. For the $5$ epochs, both the ResNet-18 and ZFNet show an MCC score of $0.71$ and  $0.69$, respectively. The model with the lowest MCC value of $0.56$ was AlexNet when trained on $25$ epochs.VGG-16 shows average results with a maximum MCC score of $0.69$ when trained for $10$ epochs.


\begin{figure}[ht] 
    \includegraphics[width=.5\textwidth]{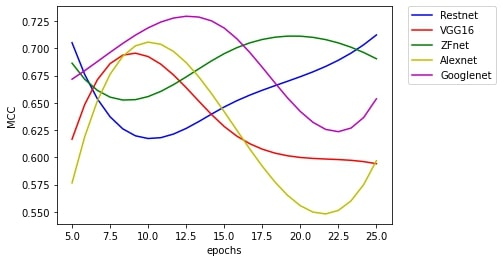}   
    \caption{MCC with augmentation.} \label{senseFNN} 
\end{figure}
The above analysis reveals that VCG-16 demonstrates the best performance in terms of sensitivity.  AlextNet and ZFNet also show better sensitivity when trained on $5$ epochs. This is due to the fact that both AlexNet and ZFNet utilizes larger filters of size $ 11 \times 11$ and $7 \times 7$, thereby recommending larger filters a better choice. Nonetheless, VGG-16 is a model that still achieves better sensitivity with small filters when trained for a higher number of epochs.  

\subsection{Performance Analysis of Transfer Learning Models without Augmentation}
All the non-augmented images are down-sampled to $224 \times 224$ before feeding it to all the selected models for transfer learning.

\begin{figure}[ht] 
    \includegraphics[width=.45\textwidth]{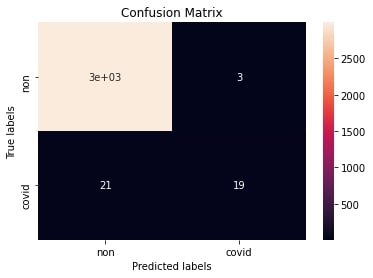}   
    \caption{ResNet-18 without Augmentation.} \label{resnet-18-noaug} 
\end{figure}

\begin{figure}[ht] 
    \includegraphics[width=.45\textwidth]{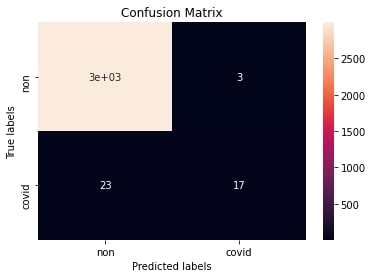}   
    \caption{ VGG-16 without Augmentation.} \label{vcg16-noaug} 
\end{figure}

\begin{figure}[ht] 
    \includegraphics[width=.45\textwidth]{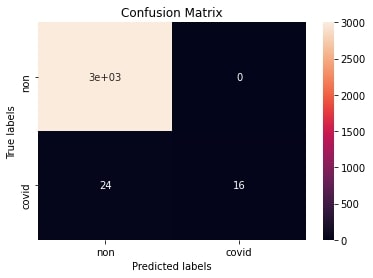}   
    \caption{ZFNet without Augmentation.} \label{zfnet-noaug} 
\end{figure}

\begin{figure}[ht] 
    \includegraphics[width=.45\textwidth]{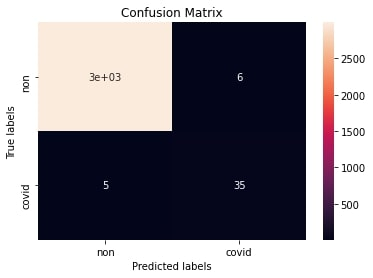}   
    \caption{AlexNet without Augmentation.} \label{alextnet-noaug} 
\end{figure}

\begin{figure}[ht] 
    \includegraphics[width=.45\textwidth]{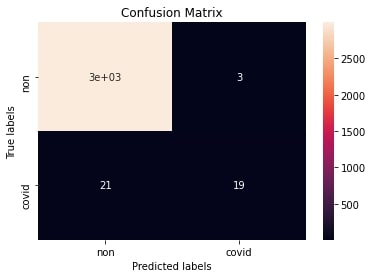}   
    \caption{ GoogleNet without Augmentation.} \label{googlenet-noaug} 
\end{figure}

Figure \ref{resnet-18-noaug} presents the confusion matrix of ResNet-18 model trained using $20$ epochs. It classifies only $19$ images as TP out of the $40$ Covid-19 positive images. In contrast to this, ResNet-18 with augmented data only classifies three images to FP. The confusion matrix of VGG-16 model trained for $15$ epochs is given in Figure \ref{vcg16-noaug}. As clear from the Figure, the model achieves $17$ TPs and $3$ FPs. Even with non-augmented data, the model tends to have a decently low number of FPs. Figure \ref{zfnet-noaug} depicts the confusion matrix of the ZFNet model trained for $25$ epochs. The model provides good results with $0$ FP for classifying health patients. On the other hand, for the infected individuals, the ZFNet underperforms and classifies only $16$ images as TP while $24$ images as FN. Figure \ref{alextnet-noaug} displays the confusion matrix of the AlexNet model trained on $25$ epochs. AlextNet shows good performance by classifying Covid-19 patients accurately, i.e., $35$ TP out of $40$ images. The transfer learned model also characterizes the Covid-19 images with good accuracy by classifying only $6$ images as FP. The confusion matrix of the  GoogleNet model trained on $15$ epochs is depicted in Figure \ref{googlenet-noaug}. The model performs well with only $3$ Covid-19 images as FP. As far as Covid-19 positive classification is concerned, the model incorrectly classifies $21$ images as FN. The model precisely classifies $19$ images as TP that is half of the total Covid-19 positive images fed into the model.


\begin{figure}[ht] 
    \includegraphics[width=.5\textwidth]{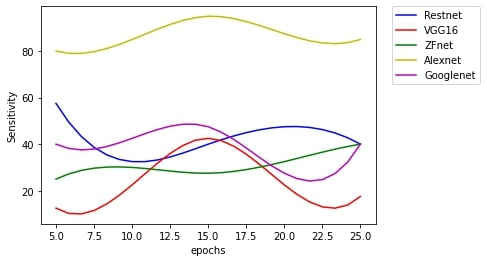}   
    \caption{Sensitivity without augmentation.} \label{sense-noaug} 
\end{figure}

Figure \ref{sense-noaug} presents the sensitivity of various Covid-19 detection models when trained on non-augmented images. For sensitivity, AlexNet demonstrates the best performance with a value of 95\% when trained using $15$ epochs. Other models show sensitivity below $60\%$ with VGG-16 with the least sensitivity when trained for $5$ epochs. When trained for $15$ epochs, AlexNet, GoogleNet, and VGG-16 achieve a sensitivity of 80\%, 47.5\%, and 42.5\%, respectively. ResNet-18 shows a sensitivity of 57.50\% when trained on $5$ epochs

\begin{figure}[ht] 
    \includegraphics[width=.5\textwidth]{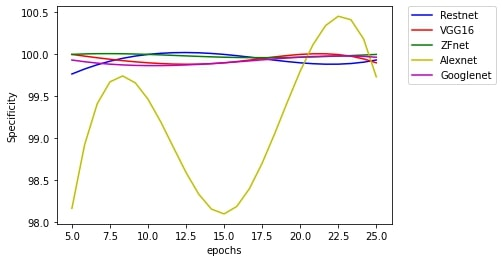}   
    \caption{Specificity without augmentation.} \label{spec-noaug} 
\end{figure}

The specificity observed by different models trained on non-augmented images is shown in Figure \ref{spec-noaug}. All the models except AlexNet demonstrate stable performance. AlexNet presents a fluctuating value of 98.17\%, 99.47\%, 98.10\%, 99.80\%, and 99.73\% when trained on $5$, $10$, $15$, $20$, and $25$ epochs.  ZFNet and VCG-16 show performance with a specificity of $100\%$ when trained on $5$ epochs.  GoogleNet and ResNet-18 also show noticeable specificity of 99.93\% and 99.77\% when trained on $5$ epochs.

\begin{figure}[ht] 
    \includegraphics[width=.5\textwidth]{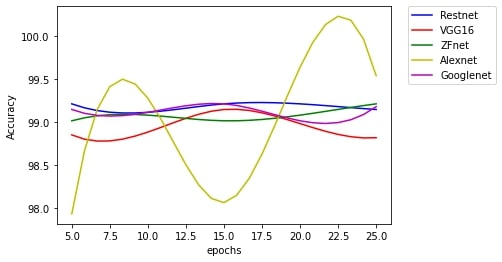}   
    \caption{Accuracy without augmentation.} \label{accu-noaug} 
\end{figure}

The accuracy of each model on non-augmented images is presented in Figure \ref{accu-noaug}. As with the specificity, all models demonstrate a stable performance with the exception of AlexNet. The accuracy demonstrated by the AlexNet is 97.93\%, 99.28\%, 98.10\%, 99.64\%, and 99.54\% when trained on $5$, $10$, $15$, $20$, and $25$ epochs. AlexNet shows the best accuracy when trained on $25$ epochs. Apart from AlexNet, ResNet-18 and GoogleNet show the best accuracy of 99.21\% and 99.14\% when trained on $5$ epochs. ZFNet and VGG-16 also show good accuracy of 99.01\% and 98.85\% when trained on $5$ epochs. 

\begin{figure}[ht] 
    \includegraphics[width=.5\textwidth]{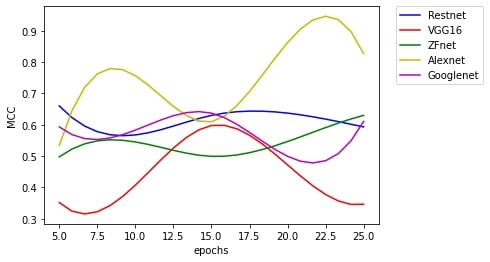}   
    \caption{MCC without augmentation.} \label{mcc-noaug} 
\end{figure}

Figure \ref{mcc-noaug} shows different values of MCC when the models are trained on non-augmented data for the Covid-19 classification. Among all the models, AlexNet achieves the highest value of 0.86 when trained on $20$ epochs. VCG-16 demonstrated an MCC value of 0.35  when trained on $5$ epochs.   ResNet-18 shows the best MCC value of 0.66 when trained for $5$ epochs. For $15$ epochs, GoogleNet, ResNet-18 and VGG-16 show a value of 0.64, 0.63, and 0.60, respectively.  Among all the models, ZFNet demonstrates a value of 0.63 when trained on $25$ epochs.

Based on experiment results, we can conclude that the best model to detect Covid-19 based on non-augmented images is AlexNet. It is mainly attributed to the size of filters used by the AlexNet. It performs better even for $5$ epochs compared to other models. As training using more epochs costs additional time and computational resources, therefore, AlexNet appears suitable with better performance with less number of epochs and with a higher sensitivity value.

Further, the experiment and evaluations reveal that most of the models perform better for Covid-19 detection when they are trained on CLAHE-augmented data. Although the specificity values of most of the models trained on non-augmented data are higher, however, most models when trained on CLAHE-augmented data achieve higher sensitivity. Sensitivity plays a significant role and is critical in determining the number of Covid-19 positive cases.  Our results suggest that most of the models demonstrate Covid-19 detection accurately with high sensitivity when X-ray images are augmented using the CLAHE method.

\section{Conclusion and Future Work} \label{conclude}

Deep learning techniques incur higher training costs due to extensive parameters tuning with a strong reliance on a large data set. Although the dataset used for model training is relatively large with 5071 Chest X-ray images consisting of both Covid-19 and normal class images \cite{van15}, the dataset remains unbalanced with only 31 images classified as Covid-19 positive for model training.  To address this issue with the data, in this work we selected various pre-trained models, including AlexNet, ZFNet, VGG-16, ResNet-18, and GoogLeNet, and transfer learned to create stable models for Covid-19 predictions. Specifically, we have demonstrated how we can augment X-ray images data by using CLAHE to train the last layer of the selected models to improve the performance of transfer learning for Covid-19 detection. Primarily, we proposed transfer learning methods based on various deep learning approaches, i.e., AlexNet, ZFNet, VGG-16, ResNet-18, and GoogLeNet.  Our results suggest that the transfer learning based on CLAHE-augmented data improves the accuracy and sensitivity of Covid-19 detection as compared to the transfer learned models based on nonaugmented data. To avoid retraining of each model, we have fine-tuned the parameters of the last layer to mitigate the possibility of over-fitting. In our experiments, we have evaluated and compared the performance of all the selected transfer learned models by using both the augmented and non-augmented data for epochs varying from $5$ to $25$. Our results reveal that the VCG-16 shows excellent performance with a sensitivity of $95\%$ when trained using $15$ epochs on CLAHE-augmented data. However, on the contrary, AlexNet works well in terms of sensitivity even when trained on non-augmented data. Other models, however, show poor performance in terms of sensitivity with a value of less than $60\%$ when trained on non-augmented data. Performance evaluations on non-augmented data in terms of sensitivity, specificity, accuracy, and MCC reveal that the sample bias can negatively impact the performance of models. In the future, we aim to use data fusion before applying CLAHE augmentation to retrain the last layer of all the selected models.

\section*{Acknowledgement}

The authors acknowledge the dataset from the Joseph Paul Cohen for the COVID-Chestxray-dataset and ChexPert dataset providers for negative data samples.

\ifCLASSOPTIONcaptionsoff
  \newpage
\fi



%

%

\begin{IEEEbiography}[{\includegraphics[width=1in,height=1.25in,clip,keepaspectratio]{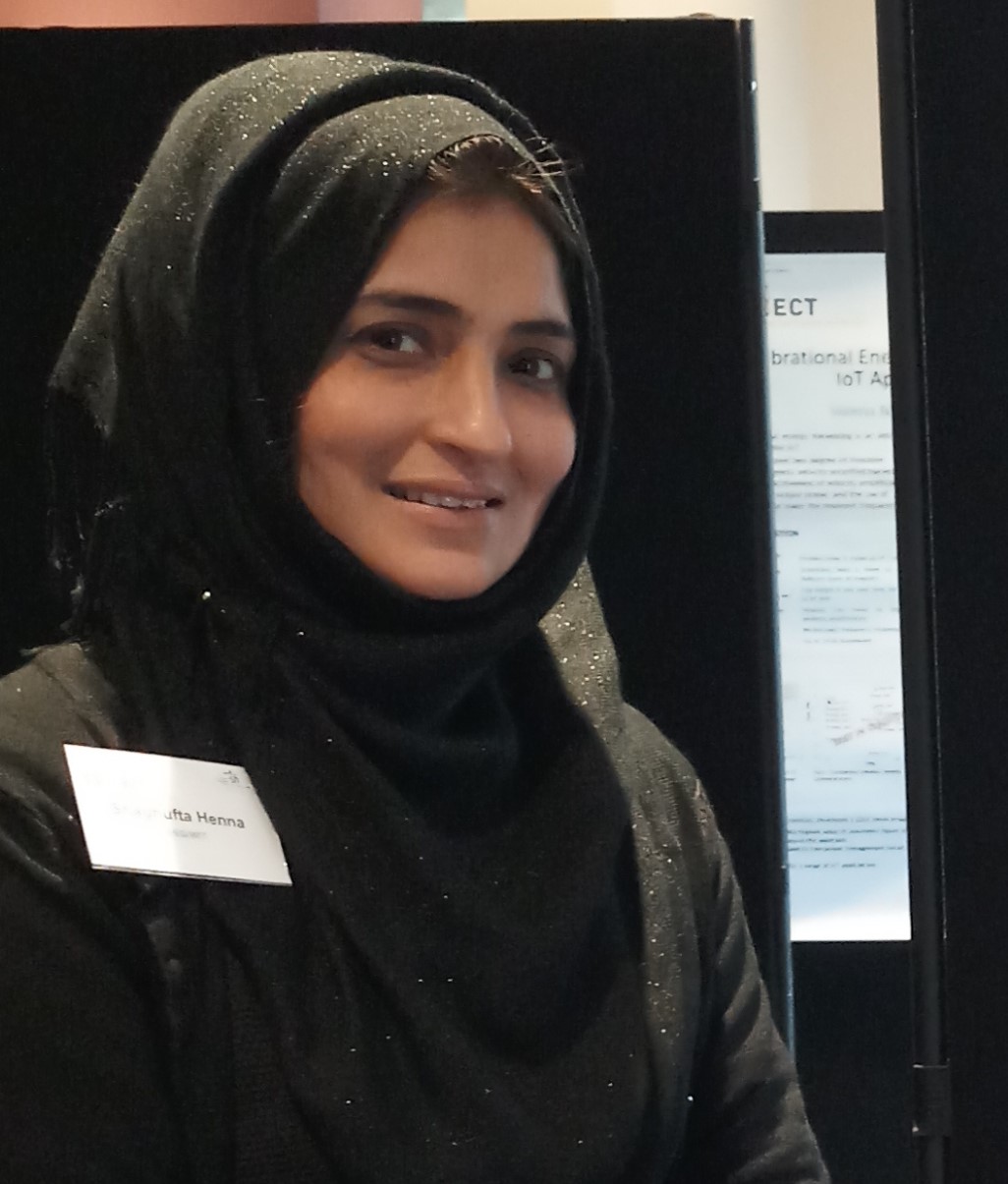}}]{Shagufta Henna} is an assistant lecturer with the Letterkenny Institute of Technology Co. Donegal, Ireland. She was a post-doctoral researcher with the telecommunication software and systems group, Waterford institute of technology, Waterford, Ireland from 2018 to 2019. She received her doctoral degree in Computer Science from the University of Leicester, UK in 2013. She is an Associate Editor for IEEE Access,EURASIP Journal on Wireless Communications and Networking, IEEE Future Directions, and Human-centric Computing and Information Sciences, Springer. Her current research
interests include big data analytics, transfer learning, distributed deep learning, and machine learning-driven network optimization.
\end{IEEEbiography}

\begin{IEEEbiography}[{\includegraphics[width=1in,height=1.25in,clip,keepaspectratio]{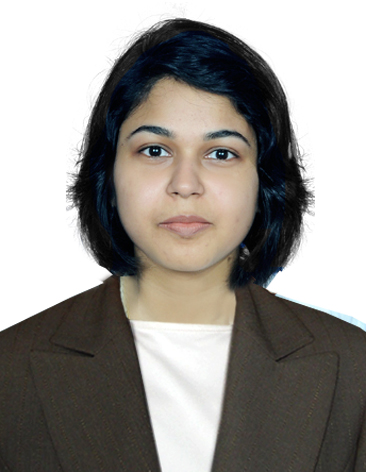}}]{Aparna Reji} is Masters student of Big Data Analytics with the Letterkenny Institute of Technology Co. Donegal, Ireland. Her current research interests include big data analytics, DevOps, and transfer learning.
\end{IEEEbiography}






\end{document}